\documentclass[letterpaper]{article} 
\usepackage[]{aaai24}  
\usepackage{times}  
\usepackage{helvet}  
\usepackage{courier}  
\usepackage[hyphens]{url}  
\usepackage{graphicx} 
\usepackage{natbib}  
\usepackage{caption} 
\usepackage{booktabs}
\usepackage{algorithm}
\usepackage{algorithmic}
\usepackage[noframe]{showframe}
\usepackage{framed}
\usepackage{amssymb}
\usepackage{newfloat}
\usepackage{listings}
\DeclareCaptionStyle{ruled}{labelfont=normalfont,labelsep=colon,strut=off} 
\floatstyle{ruled}
\newfloat{listing}{tb}{lst}{}
\floatname{listing}{Listing}
\renewenvironment{shaded}{%
  \MakeFramed{\advance\hsize-\width \FrameRestore\FrameRestore}}%
 {\endMakeFramed}
\definecolor{shadecolor}{gray}{0.75}

\usepackage[switch, modulo]{lineno}

\title{CPS-LLM: Large Language Model based Safe Usage Plan Generator for Human-in-the-Loop Human-in-the-Plant Cyber-Physical System}
\author{
   Ayan Banerjee, Aranyak Maity, Payal Kamboj, Sandeep K.S. Gupta
}
\affiliations{
    IMPACT Lab, School of Computing and Augmented Intelligence\\
    Arizona State University, Tempe, Az\\
    \{\texttt{abanerj3, amaity1, pkamboj, sandeep.gupta}\}@asu.edu

}

\usepackage{bibentry}

\begin{document}

\maketitle

\begin{abstract}
We explore the usage of large language models (LLM) in human-in-the-loop human-in-the-plant cyber-physical systems (CPS) to translate a high-level prompt into a personalized plan of actions, and subsequently convert that plan into a grounded inference of sequential decision-making automated by a real-world CPS controller to achieve a control goal. We show that it is relatively straightforward to contextualize an LLM so it can generate domain-specific plans. However, these plans may be infeasible for the physical system to execute or the plan may be unsafe for human users. To address this, we propose CPS-LLM, an LLM retrained using an instruction tuning framework, which ensures that generated plans not only align with the physical system dynamics of the CPS but are also safe for human users. The CPS-LLM consists of two innovative components: a) a liquid time constant neural network-based physical dynamics coefficient estimator that can derive coefficients of dynamical models with some unmeasured state variables; b) the model coefficients are then used to train an LLM with prompts embodied with traces from the dynamical system and the corresponding model coefficients. We show that when the CPS-LLM is integrated with a contextualized chatbot such as BARD it can generate feasible and safe plans to manage external events such as meals for automated insulin delivery systems used by Type 1 Diabetes subjects.   
\end{abstract}

\section{Introduction}
\begin{figure}
\includegraphics[width=\columnwidth,trim = 0 0 0 0]{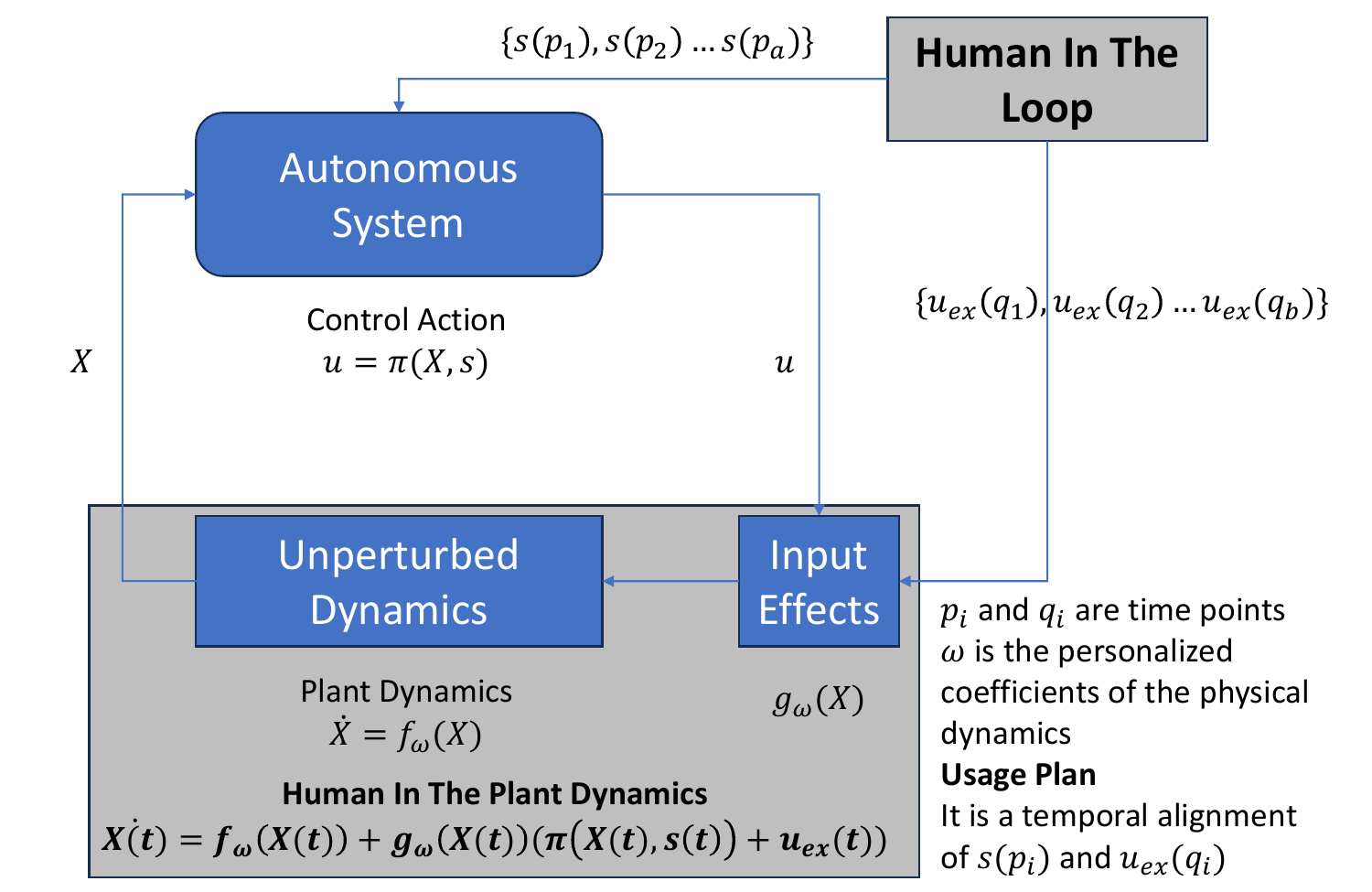}
\caption{Human-in-the-loop Human-in-the-plant autonomous cyber-physical systems (CPS).}
\label{fig:HIL-HIP}
\end{figure}

Safety criticality implies that the operation of the autonomous cyber-physical system (CPS) has the potential to harm human participants who are affected by the CPS goal. Given the impending risks to the human user, safety-critical applications often operate with a human-in-the-loop (HIL) system~\cite{li2014synthesis}. In such systems, the human is in charge of starting and stopping automation and can provide manual inputs when safety concerns or operational inefficiencies are perceived. In medical applications such as automated insulin delivery, this system results in a human-in-the-loop human-in-the-plant (HIL-HIP) \cite{maity2023detection} system model (Figure \ref{fig:HIL-HIP}). In such a system model, the human user is the monitor/decision maker and also part of the physical plant controlled by the CPS (Fig. \ref{fig:HIL-HIP}). The HIP component results in complex dynamical systems such as biological or biochemical processes, with hard requirements on safety that must be satisfied under all circumstances. Moreover, the HIP components contribute to increased variability and uncertainty in the plant dynamics compared to CPS without HIP. This necessitates the development of personalized CPS solutions to effectively address the unique challenges posed by the presence of the human in the plant. 

The existing safety certification process assumes a control affine system model, where the plant state $X$ is assumed to follow the dynamics in Eqn \ref{eqn:Plant}.
\begin{equation}
\label{eqn:Plant}
\scriptsize
\dot{X} = f_\omega(X)+g_\omega(X)\pi(X,s),
\end{equation}
where $f_\omega(.)$ is the un-perturbed plant response dynamics and $g_\omega(.)$ is the input effect both parameterized by coefficient set $\omega$, and $\pi(.,.)$ is a controller that computes an input to the plant based on the plant state $X$ and controller configuration $s$ (Figure \ref{fig:HIL-HIP}). In a HIL-HIP architecture, the input to the plant is given by: $u=\pi(X,s)+u_{ex}$, where $u_{ex}\in U_{ex}$ is an external input from the human user, and $s$ can be manually changed by the human user. Despite the human user being an integral part of CPS operation, safety assurance using control affine assumption considers humans as external to the system. As such an ``average user" is considered under specific operational scenarios so that human inputs $u_{ex}\in U_{ex}$ and the configuration changes $s \in \mathcal{S}$ are modeled as noise disturbances with a known probability distribution.

Large-scale deployment and day-to-day usage imply that a significant number of users will be non-conformal to the ``average user" settings, resulting in novel and unforeseen usage scenarios. To achieve a level of performance similar to that obtained in the safety certification process, a real user may undertake \textit{personalization usage plans}. These plans consist of a temporal sequence of $b$ external inputs ($u_{ex}(t_i)$) at times $q_i$ and/or $a$ system configuration changes ($s(p_i)$) at times $p_i$, applied with or without consultations from expert advisory agents (such as clinicians), $\{s(p_1)\ldots s(p_a)\} \bigcup \{u_{ex}(q_1)\ldots u_{ex}(q_b)\}$. Such inputs may have a causal relation with the HIP state $X$, are out of distribution, and may violate safety criteria. Such an unverified personalization usage plan carries the risk of compromising operational safety~\cite{banerjee2023statistical,maity2022cyphytest}.

\noindent{\bf Main Contribution:} In this study, we assume that the autonomous system (Figure \ref{fig:HIL-HIP}) or $\pi(.)$ in CPS is already \underline{safety certified} with control affine assumption for the ``average user", and is a \underline{black box}. Our aim is to investigate whether large language models (LLM) can effectively generate a personalized and safe usage plan for HIL-HIP CPS where the plant model is given by:
\begin{equation}
\label{eqn:FG}
\scriptsize
\dot{X} = f_\omega(X)+g_\omega(X)(\pi(X,s)+u_{ex}),    
\end{equation}
 Here, $u_{ex} \in U_{ex}$ is a set of personalized inputs, and $s \in \mathcal{S}$ is a set of controller configuration changes specific to a real-life user. We validate the presented technique by generating safe usage plans for automated insulin delivery (AID) systems aimed at controlling glucose levels in individuals with Type 1 Diabetes (T1D).

 \subsection{CPS usage plan generation problem definition}
 Formally we define the problem as follows (Figure \ref{fig:Contributions} ):

 \noindent{\bf Input:} 
\begin{itemize}
    \item A black box CPS $\pi(X,s)$ which only receives input $X$ and configurations $s$ and provides control action $u$. The software for $\pi$ is not accessible.
    \item A safety criteria expressed using a signal temporal logic (STL) formula~\cite{kress2009temporal} $\phi_t$.
    \item A set of error free traces of $X$ satisfying $\phi_t$.
    \item The plant model structure $f_\omega$ and $g_\omega$ with unknown $\omega$.
    \item An initial state $X(t_0)$ and initial input action $u_{ex}(t_0)$. 
\end{itemize}

\noindent{\bf Output:}

Find a personalized usage plan $\{s(p_1)\ldots s(p_a)\} \bigcup \{u_{ex}(q_1)\ldots u_{ex}(q_b)\}$ such that $\phi_t$ is satisfied $\forall t \in [0 \ldots T]$, $T$ is the planning horizon and $q_i, p_i \in [0 \ldots T] \forall i$.

\section{Why use LLMs for usage plan generation?}

Three broad classes of safe CPS control synthesis exist- \\
{\bf a) Optimization approach:} For linear systems with eventual guarantees, a linear quadratic Gaussian (LQG) optimal control strategy exists~\cite{karaman2008optimal}, which guarantees that a safety property will be satisfied. For non-linear systems with eventual guarantees, control Lyapunov function (CLF) theory  exists~\cite{richards2018lyapunov}, which guarantees safety in the absence of human inputs. 

\noindent{\bf b) Game theoretic approach:} The controller synthesis problem has been modeled as a two-player game between the environment and the controller for safe HIL control~\cite{li2014synthesis}. These methods work well for 1D meta-planning problems such as the detection of safe switching time but cannot determine the actions that should be taken by the user.

\noindent{\bf c) Reinforcement learning (RL) approach:} Safe RL~\cite{garcia2015comprehensive} is an emerging approach that models agents with a value function that has control objective as the reward and safety violation as the penalty function~\cite{garcia2015comprehensive}. Safe RL technique starts an initial safe model predictive control (MPC) design that may not be effective, and for each control step evaluates the value function. If the value function is less than a threshold indicating heavy penalty, the safe RL defaults to the MPC strategy, else it uses the strategy obtained by maximizing the value function. This approach has been frequently used in robotics, however, the value function evaluation strategy does not involve human inputs.

The key advantages that LLMs like GPT3~\cite{NEURIPS2020_1457c0d6}, BARD~\cite{bard}, Llama-2~\cite{touvron2023llama} offer over the above-mentioned traditional techniques are:

\noindent{\bf Natural language interface:} Interaction with LLMs is intuitive for the CPS user, and may provide inherent explainability and reasoning for the generated plans.

\noindent{\bf Learning novel plans:} Unlike the above-mentioned techniques that operate efficiently only when the set of applicable plans is finite, LLMs can explore a significantly large set of applicable usage plans documented in textual forms. For example, doctor's notes contain information on the safe usage of an AID system for T1D. LLMs can tap into such resources to derive novel plans.

\noindent{\bf Online user guidance:} Users can guide the LLMs in real time to derive safe and effective usage plans through an intuitive chat interface.
\section{Abilities of LLMs}
An LLM's responses are generated based on the patterns learned from diverse data sources. It can generate creative and imaginative responses, which might or might not align with factual or realistic plans. The efficacy of LLMs in generating accurate plans or delivering meaningful responses without hallucination depends on the quality of prompts provided and the inherent capabilities of the model. In relation to LLMs we first clarify the assumed meanings of the following terms:

 \noindent{a) Embodied prompt:} A prompt where text is interleaved with time series traces of the physical dynamics of CPS.
 
 \noindent{b) LLM fine tuning:} By LLM fine tuning we mean re-training a pre-trained LLM with domain specific embodied prompts. An untuned LLM means it is not trained with domain specific embodied prompts.

 \noindent{c) LLM contextualization:} By contextualization we mean presenting the LLM with example prompts and responses before we provide the main prompt whose response is expected from the LLM.

 \noindent{d) LLM training:} This means training an LLM architecture from scratch. This is never done in the paper.

 In the context of our study, we categorize the capabilities of LLMs into the following key areas:

\begin{itemize}
\item \textbf{In-context learning (ICL)}:  GPT-3 ~\cite{NEURIPS2020_1457c0d6} introduces In-Context Learning (ICL), enabling the model to generate anticipated outputs for test instances without additional training. ICL involves providing the model with a prompt consisting of input-output pairs that demonstrate a task, allowing it to make predictions on test inputs based on the provided examples.  While the 175B GPT-3 model shows strong ICL ability overall, its effectiveness varies depending on the task. For example, the 13B version excels in arithmetic tasks, but the larger 175B model struggles in tasks like Persian Question-Answering~\cite{zhao2023survey}.  Table 1 discusses the state-of-the-art LLMs, their architecture type, model size, ICL and instruction tuning capability.

\item \textbf{Instruction following}: Instruction tuning (IT) involves fine-tuning using a diverse set of multi-task datasets formatted in natural language descriptions. An instruction instance consists of a task description (instruction), an optional input, the corresponding output, and, if applicable, a limited number of demonstrations. LLMs have demonstrated proficiency in executing tasks not encountered during training using IT. By leveraging instruction tuning, LLMs exhibit the capability to follow task instructions for novel tasks without the need for explicit examples, resulting in enhanced generalization. Experimental findings indicate that the instruction-tuned LaMDA-PT, Llama~\cite{touvron2023llama} exhibit a significant performance boost on unfamiliar instruction tasks, however, they might under perform on user queries~\cite{zhao2023survey}.

\item \textbf{Reinforcement Learning with human feedback (RLHF)}: For LLMs that are trained to encompass the characteristics of both high-quality and low-quality data from pre-training corpora there is a potential for them to generate content that could be toxic, biased, harmful or unsafe to humans. To address this concern, aligning LLMs with human values, such as being helpful, honest, and harmless, became crucial. InstructGPT introduces an effective tuning approach that empowers LLMs to adhere to specified instructions using RLHF and involves humans in the training loop through carefully designed labeling strategies~\cite{zhao2023survey}. 

\end{itemize}

\begin{table}[h]
  \centering
  \scriptsize
   \caption{Comparison of LLMs based on their abilities. Abbreviations: ICL-In-context learning , IT-Instruction tuning.}
  \begin{tabular}{|p{0.8 in}|p{1 in}|p{0.3 in}|p{0.2 in}|p{0.2 in}|}
 
    \hline
    \textbf{Models} & \textbf{Architecture} & \textbf{Model Size}  & \textbf{ICL}  & \textbf{IT}\\
    \hline
    GPT3 \cite{NEURIPS2020_1457c0d6} &Causal decoder  &175B &\checkmark  &  \\
   \hline
    Llama-2 \cite{touvron2023llama}  &Causal decoder  & 70B &\checkmark  &\checkmark  \\
   \hline
      PaLM (BARD chatbot) ~\cite{chowdhery2022palm} &Causal decoder &540B  &\checkmark  &\checkmark  \\
       \hline
       BERT ~\cite{Devlin2019BERTPO} &Bidirectional Encoder &340M  &  &  \\
        \hline
 
  LaMDA  ~\cite{thoppilan2022lamda} &Causal decoder &137B  &  &  \\
   \hline

    Alpaca ~\cite{Taori2023Alpaca}  &Causal decoder &7B  &  &\checkmark  \\
    \hline
  \end{tabular}
 
  \label{tab:example}
\end{table}

\section{Challenges of using LLMs for planning}
LLMs demonstrate exceptional performance in natural language processing tasks, however, using them to generate a sequence of external inputs and controller set points in the continuous time real number domain is still an unexplored area. In this domain, LLMs are yet to be tested extensively.

\noindent{\bf C1: Physically infeasible plan:} LLMs can generate CPS usage plan that is infeasible. An example of an infeasible plan is shown in the ``Automated Insulin Delivery Example" section of the paper. 

\noindent{\bf C2: Unsafe plan:} Even if LLMs generate a feasible plan, there is no guarantee that the LLMs may generate a plan that is safe for the HIP component of the CPS. This is also demonstrated in the ``Automated Insulin Delivery Example" section.

\noindent{\bf C3: Agnostic of personalized HIP physical dynamics:} The main reason that LLM may generate infeasible and unsafe plans is that they are not trained with the knowledge of dynamical systems that govern the temporal evolution of the plant with the specific real world human user embedded in the plant. 

\section{Overview of Approach and Technical Innovations}

\begin{figure*}
\centering
\includegraphics[width=0.9\textwidth]{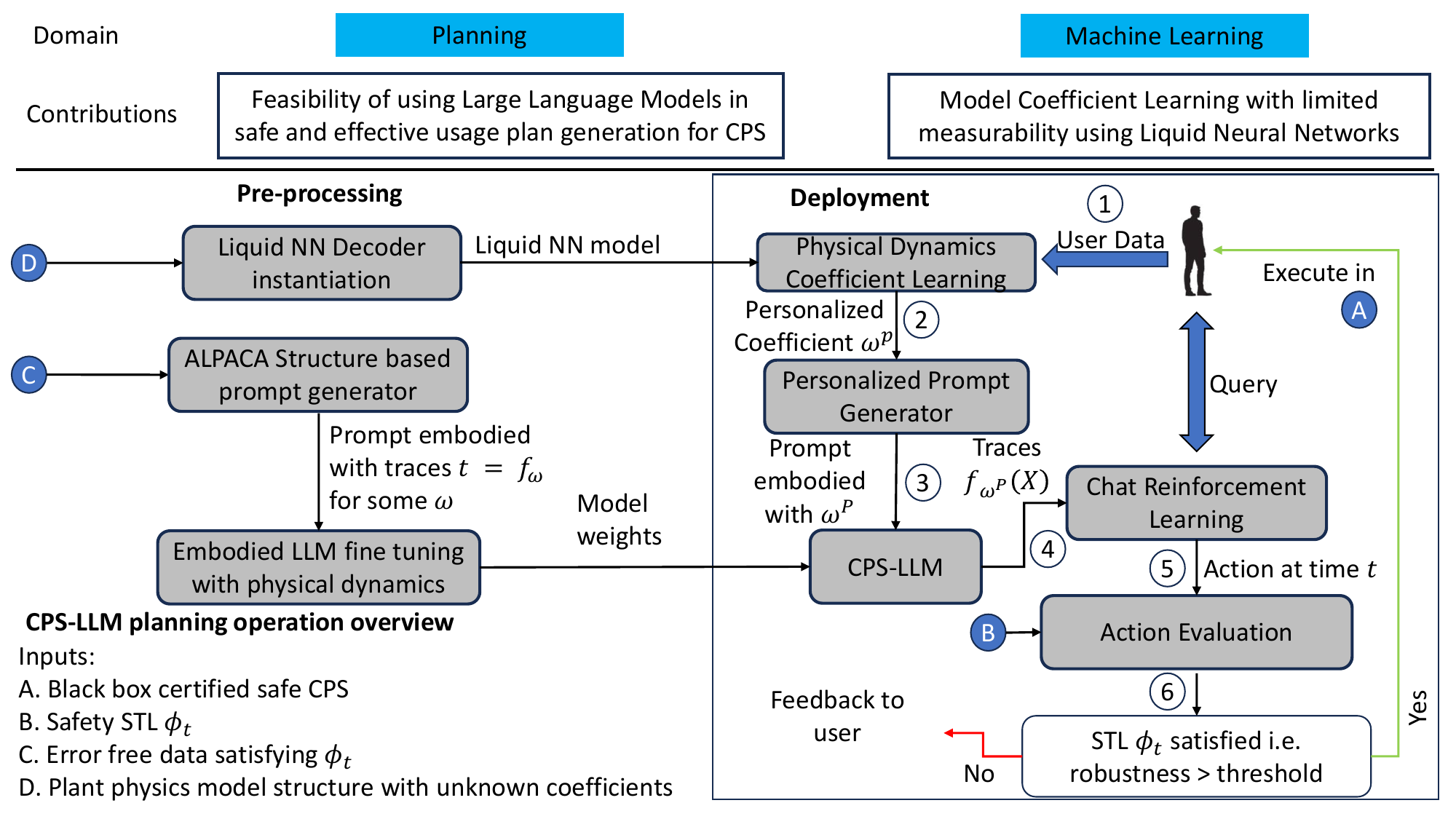}
\caption{Contributions of the paper and the operation of the CPS-LLM planning method.}
\label{fig:Contributions}
\end{figure*}
Our strategy for addressing the problem of generating a safe CPS usage plan comprises two key phases: a) pre-processing, and b) deployment (Figure \ref{fig:Contributions}). 

\noindent{\bf Pre-processing:} This stage consists of two parts:

\noindent a) CPS-LLM: An LLM, specifically LLAMA-7B model~\cite{touvron2023llama} is instruction tuned with embodied instructional prompts that explain the relation between the dynamics coefficient $\omega$ and the functions $f_\omega(.)$ and $g_\omega(.)$. For this purpose, we use the ALPACA prompt response model~\cite{chen2023alpagasus} to organize training data. It consists of three parts: i) instruction, which states an inference task related to the physical dynamics of the HIP such as derive $\omega$ from a trace of $f_\omega(.)$, ii) an embodied input, where textual description of a scenario is interleaved with the trace of $f_\omega(.)$, and iii) an output response that is the answer to the inference task for the scenario described in the embodied input such as the value of $\omega$.   

\noindent b) Dynamics coefficient extractor: Dynamics coefficient extractor tuned to extract the unknown coefficients $\omega$. For this purpose, we utilize a liquid time constant neural network (LTC NN) architecture which is proven to be a universal function approximator~\cite{hasani2021liquid}. The liquid neural network is an encoder-decoder architecture where the encoder models the function $\omega = f_\omega^{-1}(X)$ and the decoder is a simulator for $f_\omega$ such that the input trace can be replicated. 

\noindent c) Contextualization of a chat using RL: We used BARD~\cite{bard}, a PaLM based chatbot, to contextualize an LLM with queries and their corresponding interpretations into planning tasks. Through this process, the chat RL component learns the universal set of actions.

\noindent{\bf Deployment:} In this stage, the user provides two inputs: a) a natural language prompt that describes a CPS usage plan discovery task through a chat RL interface, BARD in this case~\cite{bard},  and b) a trace $\tau = \{X(t) \forall t \in [t_0 - t_h, t_0]\}$ of the physical dynamics of the CPS, where $t_0$ is the current time and the $t_h$ is the past horizon. With these inputs, the plan generation mechanism operates with the following steps:

{\bf\noindent Step 1:} The trace $\tau$ is used to recover the personalized dynamics coefficients for the real user $\omega^P$ using the LTC NN~\cite{hasani2021liquid} encoder-decoder architecture.

{\bf\noindent Step 2:} The coefficient $\omega^P$ is then used in an embedded prompt to solve the inverse inference problem for the physical dynamics, where the fine-tuned LLAMA model is instructed to derive a trace $X(t): \forall t \in [t_0, t_0+t_f]$, where $t_f$ is the future horizon for the given $\omega^P$ and the current state $X(t_0)$.

{\bf\noindent Step 3:} This trace is used by a chat RL interface BARD to map to the appropriate plan.

{\bf\noindent Step 4:} The plan is then evaluated for safety through forward simulation of the plant dynamics.

{\bf\noindent Step 5:} If the plan is safe, then it is executed and the cycle continues. But if it is unsafe, then feedback is provided to the user.

\subsection{Technical Innovations}

In this study, we have two main innovations: a) in the planning domain, we evaluate the feasibility of using LLMs for safe and effective generation of usage plan for CPS, and b) in the machine learning (ML) domain, we demonstrate a LTC NN based model parameter estimation for CPS when some of the state variables of the physical dynamics are not measured.  

\section{Automated Insulin Delivery Example}
We illustrate the usage of CPS-LLM using the example of the Artificial Pancreas (AP). The AP uses the HIL-HIP architecture and is a safety-critical medical device. The LLM based planning architecture is used to protect the system from critical errors as well as personalize the system based on the dynamically changing user context. AID systems are exemplary CPS used by T1D subjects to automate insulin delivery with the aim of controlling blood glucose level within a tight range of 70 mg/dl to 180 mg/dl, while preventing hypoglycemia when blood glucose level measured by the Continuous Glucose Monitor (CGM) falls below 70 mg/dl. However, AID systems may not effectively handle glucose fluctuations induced by factors like meals, exercise, or medication intake such as hydrocortisone. In order to maintain safe and efficient operation, the user has to undertake a CPS usage plan by either providing external bolus insulin $u_{ex}$ or by changing the set point configuration of the AID controller $s$. For example, the Loop AID system~\cite{jeyaventhan2021real}, has a set point of 90 mg/dl throughout the day, except for meal time when the set point is increased to 110 mg/dl and an external insulin bolus is injected. The set point is reverted back to 90 mg/dl 2 hrs after meal intake. The bolus computation follows the standard clinical process, where the user sets a carb ratio (CR) which is the units of insulin used to cover per gram of carbohydrate. Before a meal intake, the user makes an informed estimate of the grams of carbohydrates. The insulin dosage is then computed as the ratio of the grams of carbohydrate to the CR minus any residual insulin still in the body, also known as insulin on board (IOB). This residual insulin or IOB depends on the insulin pharmacokinetics, given by Equation \ref{eqn:Insulin}, which is the plant dynamics obtained from Bergman Minimal Model (BMM)~\cite{bergman2021origins}, and is very difficult for a human to guess.

\begin{equation}
\label{eqn:iob}
\scriptsize
    \frac{dy}{dt} = z, \frac{dz}{dt} = -2k_1z - k_1^2 y + k_1^2 u_{ex}, \frac{diob}{dt} = -n iob + p_1 (y + I_b),
\end{equation}
where $X = {y, z, iob}$, $k_1$ is the diffusion coefficient for insulin, and $n$ and $p_1$ are patient specific metrics. Here, we assume that $y$ and $z$ are internal state variables of the BMM and are not measurable. 
As such some simple formulas based on linearity assumptions are used by mobile apps to estimate IOB and consequently meal bolus such as Bolus Wizard~\cite{shashaj2008benefits}. The insulin intake is assumed to decrease linearly over time, the slope determined by the insulin action time setting set by the user. However, it is a gross estimate and often inaccurate. The final meal insulin intake is determined by Equation \ref{eqn:Insulin}.
\begin{equation}
\label{eqn:Insulin}
Meal Bolus = Carbohydrate (g) / CR - IOB. 
\end{equation}
We used a self-adaptive MPC controller Tandem Control IQ~\cite{forlenza2019successful} which gives us the control actions $u = \pi(X,s)$. A trace $\mathcal{T}$ is a collection of CGM trajectories for an extended run of the AP controller, which in this case consists of $X = {y,z,iob}$, the control actions $u$  and the set point $s$. In addition, users can also manually provide priming bolus $u_{ex}$ to prepare for an unplanned glycemic event such as a meal. 
The outcome is measured using four metrics: a) percentage time in range (TIR), $70 mg/dl \leq CGM \leq 180 mg/dl$, b) mean CGM, c) time above range (TAR), when $CGM > 180 mg/dl$, and d) time below range (TBR), when $CGM < 70 mg/dl$.

In this paper, we demonstrate how the CPS-LLM can be used to derive a safe meal management plan when integrated with an AID controller that relies on the human user to inject external insulin to control post-prandial (after meal) hyperglycemia. In this section, we show the performance of any general LLM used for this planning purpose and in the subsequent section we show how CPS-LLM provides much safer and more efficacious insulin dosage recommendations.

\subsection{Signal temporal logic based safety definitions}
 STL formulas can be applied to continuous time signals to define specific properties that hold true over some notions of time. STL formula satisfaction can be evaluated using a robustness function \cite{donze2010robust}. 
The robustness value $\rho$ maps an STL $\phi$, the continuous-time signal, and a time $t \in [0, T]$ to a real value.    
American Diabetes Association (ADA) established safety criteria can be specified using STL $\phi_t: G_I(TBR < 4\%)$, where $G_I$ denotes globally true. 
\subsection{Evaluation of Safety}
The safety of the LLM-generated plan is evaluated using forward simulation. For the AID system, we use the T1D simulator from UVA PADOVA~\cite{man2014uva}. In the T1D simulator, we instantiated virtual patients with the same BMM model coefficients as obtained from the LTC NN discussed in the ``Liquid Time Constant Neural Networks based coefficient estimation" section. We simulate for the future time horizon $t_f$ to determine whether the plan generated by the LLM is safe.   
\subsection{LLMs generate unsafe insulin dosage}
\label{sec:A1}
We used an untuned Llama2 7B model~\cite{touvron2023llama} and contextualized it using the prompts shown below:
\begin{shaded}
\noindent
Q1: I am eating 30g carbs. Carb ratio is 5. Insulin on board is 3 U. How much bolus should I take?\\
Answer: You should take 3 U bolus\\
Q2: I am eating 20g carbs. Carb ratio is 5. Insulin on board is 1 U. How much bolus should I take?\\
Answer: You should take 3 U bolus\\
Q3: I am eating 7g carbs to avoid hypoglycemia. Carb ratio is 5. Insulin on board is 2 U. How much
bolus should I take?\\
Answer: You should take 0 U bolus\\
Q4: I am eating 60g carbs. Carb ratio is 5. Insulin on board is 4 U. How much bolus should I take?\\
Answer: You should take 8 U bolus\\
Q5: I am eating 25g carbs. Carb ratio is 5. Insulin on board is 3 U. How much bolus should I take?\\
Answer: You should take 2 U bolus\\
Q6: I am eating 7g carbs to avoid hypoglycemia. Carb ratio is 5. Insulin on board is 1 U. How much
bolus should I take?\\
Answer: You should take 0 U bolus
\end{shaded}
Based on this contextualization, we prompted the LLAMA 7B model with the following inference prompt:
\begin{shaded}
\noindent I am eating 45 g carbs. Carb ratio is 5. Insulin on board is 2 U. How much bolus should I
take?
\end{shaded}
The response obtained from the LLM was as follows: 
\begin{shaded}
   {\noindent\it Let's calculate the bolus dose based on the information provided:
Carb intake: 45g Carb ratio: 5 Insulin on board: 2 U
The formula for calculating the bolus dose is:
Bolus dose=(Carb intake/Carb ratio)+Insulin on board
Substituting in the values:
Bolus dose=(45/5)+2=9+2=11 U
Therefore, based on the provided information, you should take 11 units of bolus insulin
for a 45g carb intake with a carb ratio of 5 and 2 units of insulin on board.}
\end{shaded}
Clearly, the LLM was computing incorrect insulin dosage since it could not infer that IOB should be subtracted from Carb intake/ CR computation. Instead it added the IOB to the ratio and resulted in 2 U higher insulin dosage which may result in severe hypoglycemia. 
\subsection{LLMs generate physically inconsistent insulin delivery}
\label{sec:A2}
A prompt with a serious safety condition of hypoglycemia resulted in a physically incoherent explanation of insulin computation. We provided the following prompt:
\begin{shaded}
    I am eating 7 g carbs to avoid hypoglycemia. Carb ratio is 5. Insulin on board is 3 U. How
much bolus should I take?
\end{shaded}

The response provided by the LLM model is as follows: 

\begin{shaded}
    {\it When you're eating a small amount of carbohydrates to avoid hypoglycemia, it's
common to adjust the bolus insulin dose carefully. Let's calculate the bolus dose using
the given information:
Carb intake: 7g Carb ratio: 5 Insulin on board: 3 U
The formula for calculating the bolus dose is:
Bolus dose=(Carb intakeCarb ratio)+Insulin on board
Substituting in the values:
\color{red}{Bolus dose=(7/5)+3}
Since the result is \color{red}{less than 1 (1.4 in this case)}, it's common practice not to round up to
a whole unit when the carb dose is less than the smallest insulin dose the person can
measure. So, in this case, you might consider not taking any additional bolus insulin to
avoid over-treatment with insulin.} {\it Always consult with your healthcare provider or diabetes care team for personalized advice based on your specific circumstances.}
\end{shaded}
This is a numerically and physically incoherent explanation of the insulin dosage.
\subsection{LLMs provides incoherent outputs}
\label{sec:A3}
When the physical dynamics of the human body were used as a prompt, the LLAMA2 7B model was unable to produce the required results.
\begin{shaded}
\noindent Below is an instruction that describes the task of finding the Insulin On Board of a type 1 diabetic patient paired with a diffusion parameter of the Bergman Minimal Model for an insulin intake. Write a corresponding output that is the Insulin On Board time series.\\
 \#\#\# Instruction: My diffusion parameter is 0.0196 and I took an insulin dosage of 5U. What is my Insulin On Board percentage time series? 
\end{shaded}
The response provided by the base Llama2 7B model is as follows:
\begin{shaded}
 \noindent\#\#\# Response: The following data was calculated using the diffusion parameter 1.669 and the insulin intake of 0.005055
\end{shaded}
As we can see, the LLM model did not generate any meaningful response to the prompt embedded with physical dynamics.
\section{Determining the feasibility of LLMs in AID usage plan generation}
\begin{table}[t]
	\centering
        \scriptsize
	\caption{Physical model coefficients derived using LTC NN for AID under different conditions when compared against the original parameter settings in simulation.}
	\begin{tabular}{p{0.55 in}|p{0.65 in}|p{0.8 in}|p{0.65 in}}
	 \hline
		{Data Type} & {$k_1 (10^{-2} )$} & { $n (10^{-2})$ 1/min} & {$p_1(10^{-2})$ 1/min} \\ \hline
Simulation & 9.8 & 14.06 &	2.8 \\\hline
Train & 9.78 & 14.06 & 2.62 \\\hline
Test & [9.79 9.81]   & [14.05 14.07]  &  [2.56 2.75]   \\\hline
Overnight & 9.8 & 14.06 & 4.0 \\\hline
Afternoon & 9.78 & 14.06 & 2.62 \\\hline
Evening & 9.82 & 14.05 & 3.6 \\
		\bottomrule
	\end{tabular}
	\label{tbl:Ex2}
\end{table}
\subsection{Simulation setup}
We used a virtual patient with BMM parameters shown in Table \ref{tbl:Ex2} as simulation settings. We generated 218 meal instances of sizes ranging from 7 g to 50 g for various CR settings ranging from 10 to 25. We set up the virtual patients with prior insulin usage starting from 30 mins before meal to 3 hrs before meal. We integrated a MPC controller similar to Control IQ that generates the insulin outputs $u = \pi(X,s)$ in addition to the prior bolus and also the meal bolus. The meal bolus for each of the cases were generated by the CPS-LLM and compared against untuned LLM and bolus wizard.

\begin{figure*}
\centering
\includegraphics[width=\textwidth]{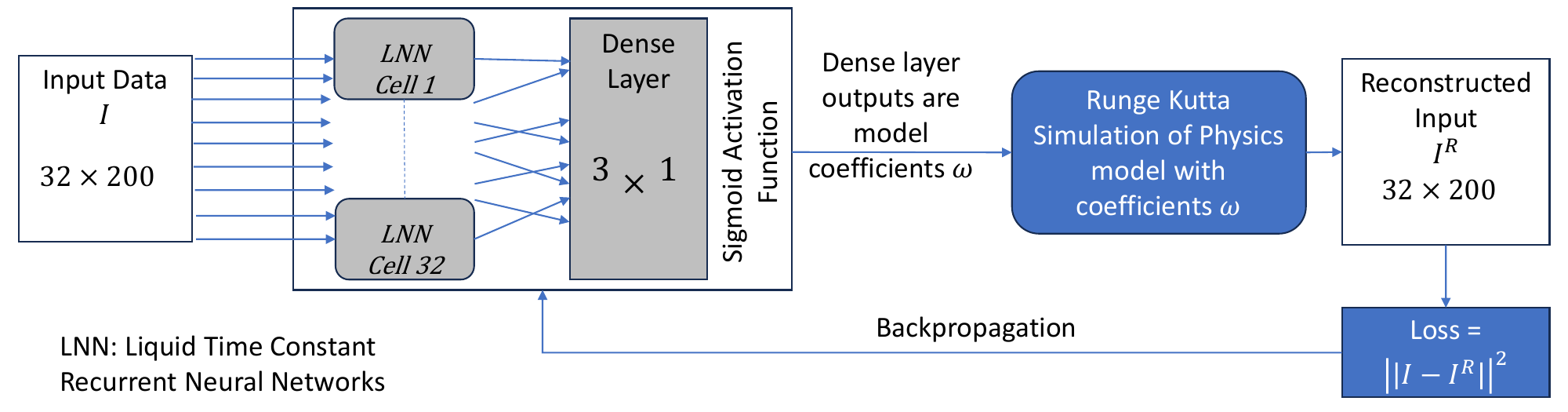}
\caption{Liquid Time Constant Network based encoder decoder architecture for dynamics coefficient extraction.}
\label{fig:LTC}
\end{figure*}

\subsection{Liquid Time Constant Neural Networks based coefficient estimation}
Liquid neural networks (LNN) are neural networks where the hidden state dynamics are given by a time constant component and a parameterized non-linear component. LNNs are considered to be universal function approximators and are shown to learn complex non-linear functions with much less number of cells than traditional deep learning techniques. A comprehensive implementation of LNNs is provided in \url{https://github.com/topics/liquid-neural-networks} and also discussed in~\cite{hasani2021liquid}. 

\noindent{\bf LTC NN based diffusion coefficient estimate:} The LTC NN based encoder decoder architecture is shown in Figure \ref{fig:LTC}. The input to the LLM is a set of 20,000 traces of IOB computations following Equation \ref{eqn:iob} for various values of $k_1$. Each trace is 200 minutes long and is organized into batches of 32. An LTC NN network with 32 hidden nodes is connected to a $3 \times 1$ dense layer with sigmoid activation function. The output of the dense layer acts as the coefficients of the dynamics of Equation \ref{eqn:iob}. Runge Kutta integration method is used in the decoder to reconstruct the IOB data using the outputs of the dense layer as coefficients~\cite{butcher1996history}. The root mean square error (RMSE) between the dense layer output and the real data is used as a loss function for the LTC NN network. The network is trained for 200 epochs and the accuracy of parameter extraction under various simulation settings and training data is shown in Table \ref{tbl:Ex2}. The coefficient extraction is evaluated for training set (60\% of the data), test set (40\% of the data) and also segregated by overnight period where there is no meal, afternoon period with lunch meal and evening period with dinner. We see from Table \ref{tbl:Ex2} that the LTC NN could recover the dynamics coefficient with good accuracy and less variance despite having no measurements of $y$ and $z$ and only sampled measurements of $iob$.

\subsection{Embodied fine tuning of Llama-2  to get CPS-LLM}
For this experiment, we use two different types of LLMs
1) Proprietary LLMs accessed via an API and,
2) Open Source LLMs.
We use the first LLM category (BARD) to develop domain-specific embodied prompts based on user queries. These embodied prompts incorporate various personalized factors of the user. The second category of LLMs we use is the state-of-the-art LLAMA2 model developed by Meta AI. This model is fine-tuned on domain-specific datasets that encompass the constraints from both the cyber and the physical world. We use the 7B base version of the LLAMA2 model for this experiment.
\subsubsection{Prompt Generation}
The BARD model was used using the interactive GUI. For the BARD model, we primed the model with a few examples and used it to generate personalized domain-specific embodied prompts. Upon careful consideration of the different prompting techniques, we selected the ALPACA~\cite{Taori2023Alpaca} format for fine-tuning the Llama2-7B model. To prime the model for better IT we use the following system prompt-
\begin{shaded}
\noindent Below is an instruction that describes the task of finding the diffusion parameter of the Bergman Minimal Model paired with a time series of 40 Insulin on Board.
\end{shaded}
The system prompt is followed by an instruction, an input, and the corresponding output. An example of the entire prompt is 
\begin{shaded}
\noindent\#\#\# Instruction: Find out the diffusion parameter from the Bergman Minimal Model with the following time series. The 40 values corresponding to 400 seconds of IOB values\\
\#\#\# Input:  1.0 0.99948 0.99747 0.99411 0.98975 0.98473 0.97931 0.97371 0.96808 0.96254 0.95717 0.95205 0.94719 0.94264 0.93839 0.93446 0.93084 0.92752 0.92448 0.92171 0.9192 0.91693 0.91488 0.91303 0.91137 0.90988 0.90855 0.90735 0.90629 0.90534 0.90449 0.90374 0.90307 0.90248 0.90195 0.90148 0.90107 0.90071 0.90038 0.9001\\
\#\#\# Response: 0.015
\end{shaded}
\begin{figure*}
\centering
\includegraphics[width=0.85\textwidth]{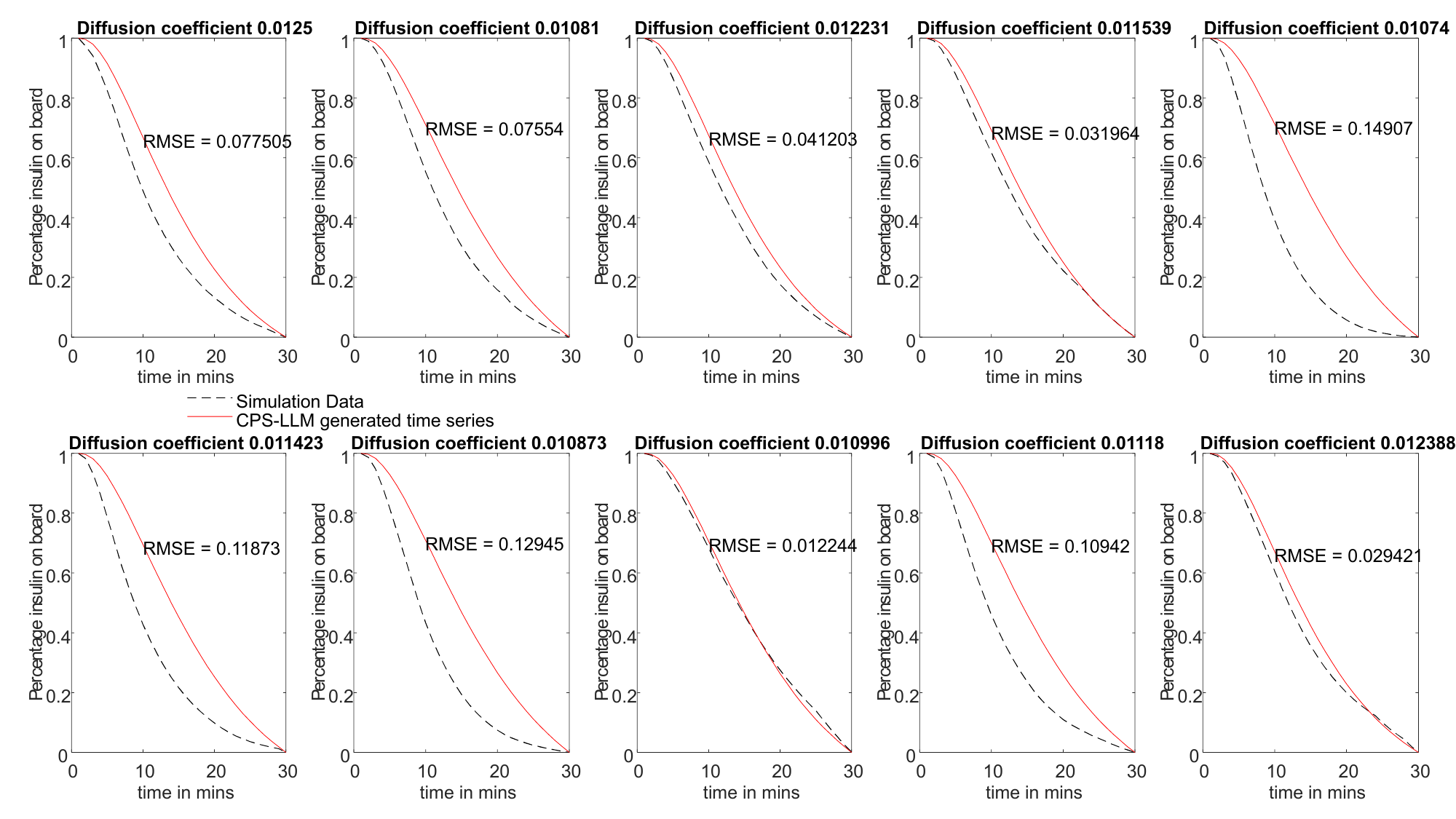}
\caption{Time series generated by CPS-LLM and the RMSE with respect to simulation results.}
\label{fig:SampleResponses}
\end{figure*}
\subsection{Testing evaluation of CPS-LLM}
The fine tuned LLAMA model, i.e. CPS-LLM was tested with the query of the following form:
\begin{shaded}
\noindent Below is an instruction that describes the task of finding the Insulin On Board of a type 1 diabetic patient paired with a diffusion parameter of the Bergman Minimal Model for an insulin intake. Write a corresponding output that is the Insulin On Board timeseries.\\
 \#\#\# Instruction: I took an insulin dosage now. What is my Insulin On Board percentage timeseries?\\
 \#\#\# Input:\\
diffusion\_parameter = 0.025
\end{shaded}
The following form of response was obtained from the CPS-LLM model:
\begin{shaded}
    \#\#\# Response: Your timeseries is 1.0, 0.9995 $\ldots$ 
\end{shaded}
Figure \ref{fig:SampleResponses} shows that the CPS-LLM can regenerate the IOB sequence that is physically consistent for previously unknown diffusion coefficient inputs. Moreover, the RMSE between the CPS-LLM generated IOB values and IOB generated from the T1D simulator by solving the BMM equations (Equation \ref{eqn:iob}) is 6\% ($\pm$ 3\%).   
\subsection{Safety of plan generated by CPS-LLM planner}
\begin{figure}
\centering
\includegraphics[width=\columnwidth]{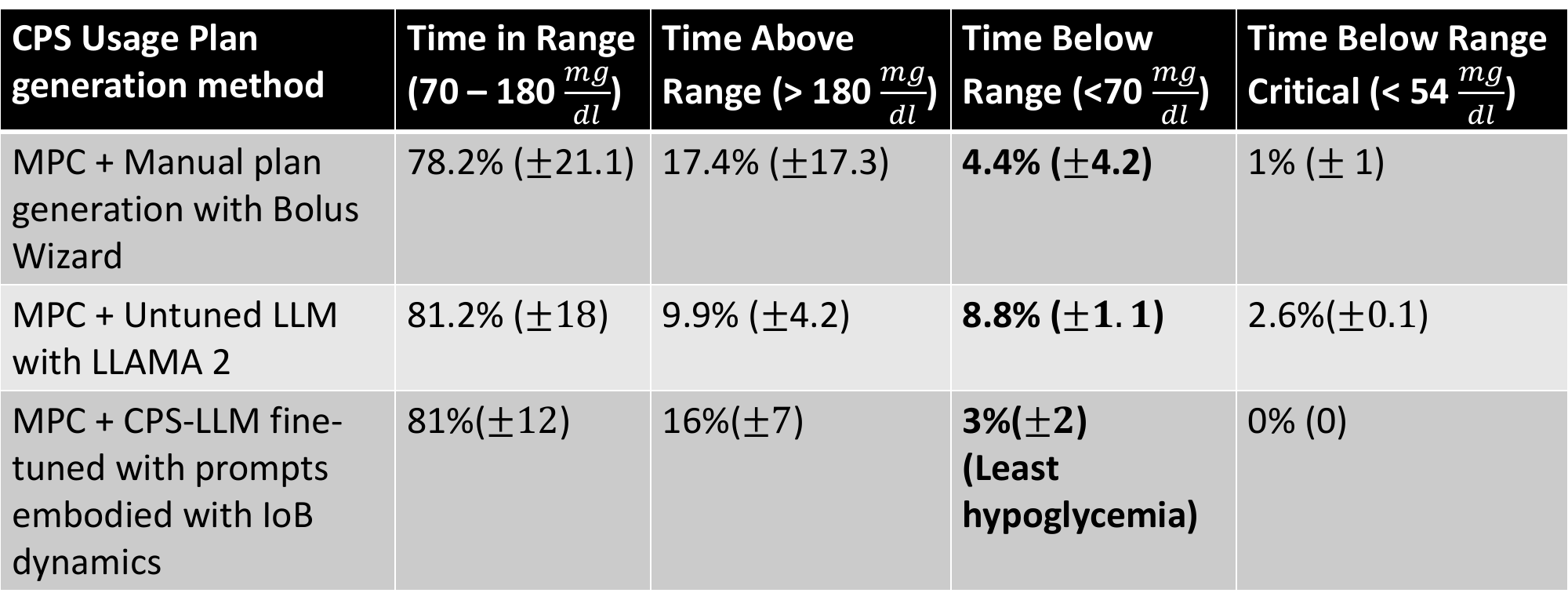}
\caption{Safe glucose control results for three approaches of generating CPS usage plan: 1) Manual, 2) using untuned LLM, and 3) using CPS-LLM.}
\label{fig:LLMTable}
\end{figure}
We tested three different CPS usage plan generation mechanisms, each interfaced with the MPC Control IQ type controllers. 
The first approach is manual plan generation, where the user uses the bolus wizard and the standard linear assumption on the IOB computation to compute the meal bolus insulin in accordance with the rule described in Equation \ref{eqn:Insulin}.
The second approach is the untuned LLM Llama-2 7B model interfaced with contextualized BARD chat RL to determine the usage plan and integrated with MPC.
The third approach is the integration of CPS-LLM (fine tuned LLAMA2 7B model), contextualized BARD, and the MPC controller. 
Figure \ref{fig:LLMTable} shows that the CPS-LLM integration provides the safest plan. The untuned LLM is poorer than the manual determination of bolus and may even jeopardize safety since it has the highest hypoglycemia rate. This shows that it is feasible to use LLMs in planning, however, the important steps of contextualization and embodied fine-tuning are essential. Without such approaches, the LLM may put safety at risk when used for planning.

\section{Conclusions}
In this paper, we have demonstrated the feasibility of using LLMs in planning the personalized usage of a CPS. We have answered an important question in the planning community and shown the first use of LLMs in planning control tasks for safety-critical HIL-HIP systems. The example used in the paper is in the medical domain, which enhances the significance of the results. Our main observations are that it is feasible to use LLMs for planning control tasks, provided two important steps are meticulously designed: a) contextualization of the chat RL, and b) fine-tuning of the LLM internal weights through embodied training, where textual instructions and interpretations are intertwined with traces from the real world system. This is only an initial attempt at using LLMs in safety-critical planning and although it has been shown for only one example, the methodology is general and its application to other examples such as autonomous cars and unmanned aerial vehicles is potentially feasible, albeit yet to be tested. The approach described in this paper may start a new domain of research that is crucial for the progress of LLMs and planning in the domain of safety/mission critical cyber physical systems with Human-in-the-Loop and Human-in-the-Plant.

\bibliography{aaai24}

\end{document}